\documentclass[11pt,a4paper]{article}
\usepackage[hyperref]{acl2019}
\usepackage{times}
\usepackage{latexsym}
\usepackage{verbatim}
\usepackage{xspace}
\usepackage{booktabs}
\usepackage{graphicx}

\usepackage{url}

\aclfinalcopy

\newcommand{\std}[1]{\textcolor{gray}{(#1)}}
\newcommand*{\al}[2]{\multicolumn{1}{#1}{#2}}
\newcommand{\tabitem}{~~\llap{\textbullet}~~}

\newcommand{\dialogue}{\textsc{Dialogue}\xspace}
\newcommand{\satisfaction}{\textsc{Satisfaction}\xspace}
\newcommand{\feedback}{\textsc{Feedback}\xspace}
\newcommand{\personachat}{\textsc{PersonaChat}\xspace}

\title{Learning from Dialogue after Deployment:\\
Feed Yourself, Chatbot!}

\author{
  Braden Hancock\thanks{*BH completed most of this work at Facebook (FAIR).} \qquad \quad \thinspace \\
  Computer Science Dept. \qquad \quad \thinspace \\
  Stanford University \qquad \quad \thinspace \\
  {\tt bradenjh@cs.stanford.edu} \qquad \quad \thinspace
  \And
  \qquad \quad Antoine Bordes, Pierre-Emmanuel Mazar\'{e} \\
  \qquad \quad \textbf{Jason Weston} \\
  \qquad \quad Facebook AI Research \\
  \qquad \quad {\tt \{abordes,pem,jase\}@fb.com}
}

\date{}

\begin{document}
\maketitle

\begin{abstract}
The majority of conversations a dialogue agent sees over its lifetime occur after it has already been trained and deployed, leaving a vast store of potential training signal untapped.
In this work, we propose the self-feeding chatbot, a dialogue agent with the ability to extract new training examples from the conversations it participates in.
As our agent engages in conversation, it also estimates user satisfaction in its responses.
When the conversation appears to be going well, the user's responses become new training examples to imitate.
When the agent believes it has made a mistake, it asks for feedback; learning to predict the feedback that will be given improves the chatbot's dialogue abilities further.
On the \personachat chit-chat dataset with over 131k training examples, we find that  learning from dialogue
with a self-feeding chatbot significantly improves performance, regardless of the amount of %
traditional supervision.
\end{abstract}

\begin{figure}[t]
    \centering
    \includegraphics[width=3in]{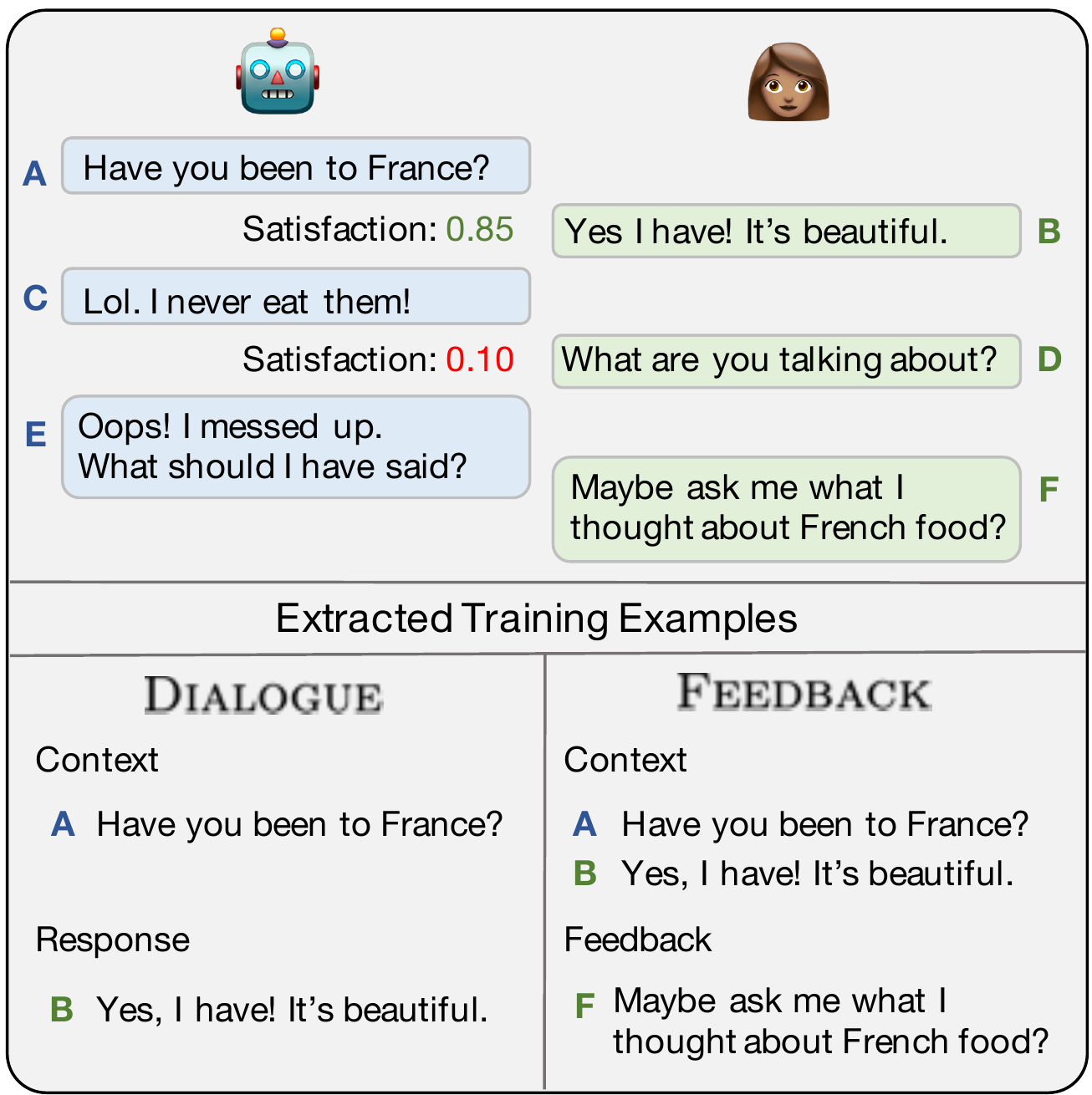}
    \caption{
        As the self-feeding chatbot engages in dialogue, it estimates user satisfaction to know when to ask for feedback.
        From the satisfied responses and feedback responses, new training examples are extracted for the \dialogue and \feedback tasks, respectively, both of which improve the model's dialogue abilities further. 
    }
    \label{fig:interface}
\end{figure}

\section{Introduction}
Training a dialogue agent to converse like a human requires extensive supervision.
The most common approach is to train models to imitate humans in large corpora of crowdsourced or scraped conversations \cite{serban2015survey}.
These fully-supervised conversations tend to be expensive to collect in sufficient quantity and/or occur in settings with significant differences from the deployment environment \cite{ross2009turkers}.
Instead, dialogue agents would ideally learn directly from dialogue, the conversations they participate in after deployment, which are usually abundant, task-specific, dynamic, and cheap.
This corresponds to the way humans learn to converse---not merely observing others engaging in ``expert-level'' conversations, but instead actively adjusting and correcting our speech based on feedback woven throughout our own conversations \cite{bassiri2011interactional,werts1995instructive}.
Giving a dialogue agent this ability would enable it to continuously improve and adapt over its lifetime, rather than requiring additional annotation costs for each and every improvement.

However, naively training a dialogue agent on its own conversations yields poor results.
For example, training a model on its own output can simply reinforce its existing failure modes, and mistakes by the agent can lead to absurd conversations that no longer resemble the target domain \cite{hashimoto2018detecting}.
To combat this, one approach is to allow the agent to request feedback during conversations \citep{zhang2018interactive,li2017learning}, e.g., when it believes it is about to make a mistake.
This approach, however, falls victim to the Dunning-Kruger effect \citep{kruger1999unskilled}, which in this case suggests that a bad model will also be bad at knowing when it is doing a bad job.
Regardless of when feedback is requested, existing methods typically require accompanying scalar rewards or adherence to particular templates or structure to ensure that the feedback is usable by the model \citep{rieser2011reinforcement,zhang2017listen,liu2018dialogue}.
These requirements may be acceptable for paid annotators, but they impose unnatural workflows on unpaid conversation partners in a standard dialogue environment.
Humans are able to request and provide feedback using only natural language; ideally, dialogue agents would be able to do the same.

In this work we propose the self-feeding chatbot, a dialogue agent with the ability to extract new examples from the conversations it participates in after deployment (Figure~\ref{fig:interface}).
Concretely, in addition to being trained on the primary \dialogue task, the agent is trained to predict its speaking partner's satisfaction with its responses.
When the conversation seems to be going well, the user's responses (but not the bot's own utterances) become the targets in new training examples for the \dialogue task.
When the agent believes it has made a mistake, it instead requests feedback on what it could have said instead.
Predicting the feedback that will be provided in a given context becomes an auxiliary task (\feedback) on which the model is also trained.
Importantly, these new examples improve the agent's dialogue abilities while using only natural responses from the user that do not require special structure, accompanying numerical feedback, or additional human intervention in order to be used.

With this approach, the conversations the chatbot participates in are sliced into two complementary datasets---one largely protected from the chatbot's mistakes (\dialogue examples), and one which directly addresses them (\feedback examples).
We validate our approach on the \personachat \citep{zhang2018personalizing} dialogue dataset, finding empirically that regardless of the number of available supervised examples, the dialogue ability of the chatbot is always improved by adding the automatically extracted examples of either type, and improves the most by adding both.

The main contributions of this work thus include the following:
\begin{itemize}
    \item{We propose the self-feeding chatbot, a dialogue agent with the ability to extract new training examples for itself from the conversations it participates in during deployment.}
    \item{We show that dialogue ability improves by imitating human responses when the human is satisfied, or by asking for feedback when they are not, predicting it as an auxiliary task.
    }
    \item{We demonstrate that classifying user satisfaction is a learnable task important for the self-feeding process, significantly outperforming an approach based on model uncertainty.}
    \item{We release three new datasets to further research in this direction: (1) deployment chat logs (513k messages); (2)
    ratings of user satisfaction (42k); (3) textual feedback on what a bot could have said in a given context (62k).}
\end{itemize}

\noindent
The datasets and models described in this paper are available via the ParlAI platform \citep{miller2017parlai}, along with training code.
Hyperparameter values are included in Appendix~\ref{app:hyperparameters}.

\begin{figure*}[t]
    \centering
    \includegraphics[width=6in]{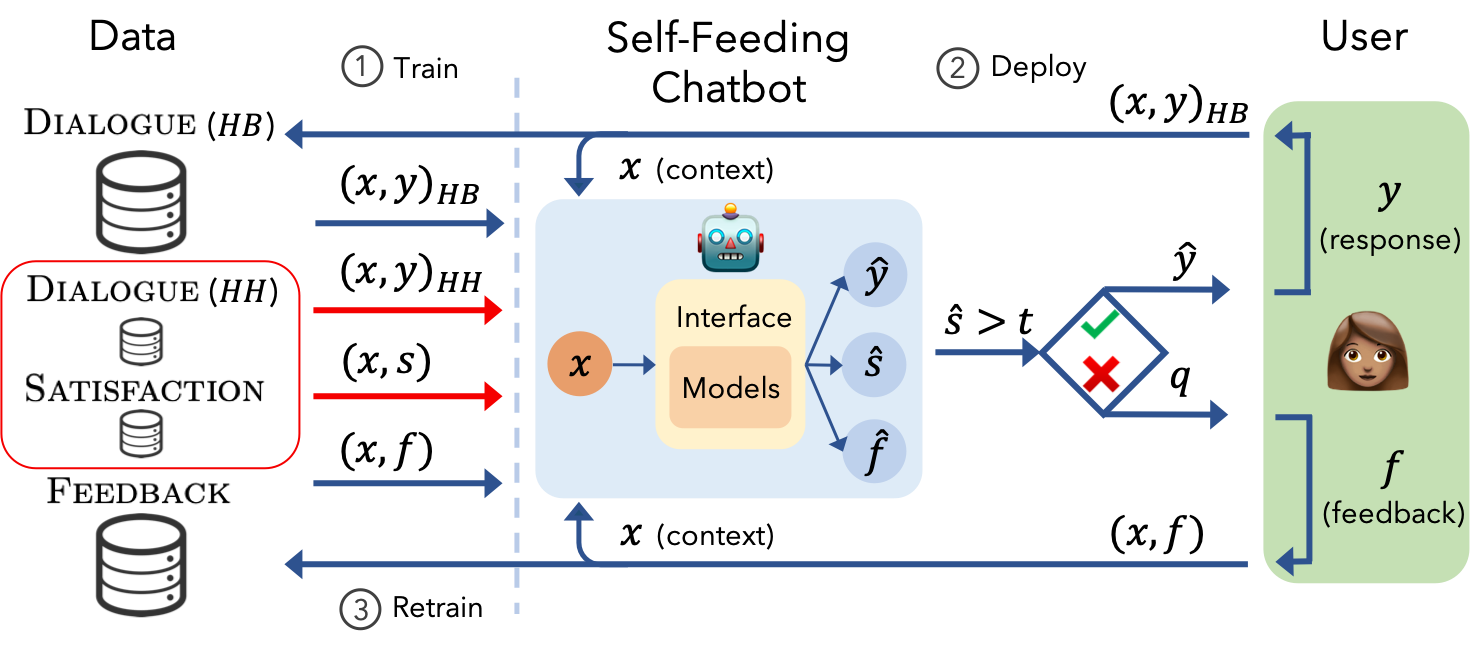}
    \caption{(1) The chatbot is first trained with any available supervised data (boxed in red) on the Human-Human (HH) \dialogue $(x,y)_{HH}$ and \satisfaction $(x,s)$ tasks. (2) During deployment, whenever the predicted satisfaction score of the current conversation $x$ is above the threshold ($\hat{s} > t$), a new Human-Bot (HB) \dialogue example $(x,y)_{HB}$ is extracted %
    and the bot continues the conversation with its own response $\hat{y}$. Otherwise, the chatbot requests feedback with question $q$ and extracts a new \feedback example $(x,f)$. (3) The chatbot is periodically retrained with the available examples from all four datasets, improving its \dialogue performance without collecting any new supervised examples.}
    \label{fig:banner}
  \end{figure*}

\section{Related Work}
\label{sec:related}

The general concepts of lifelong learning \citep{silver2013lifelong} and 
never-ending (language) learning \cite{carlson2010toward} are related to the topics discussed in this work, as is active learning
\cite{tong2001support} 
and predictive modeling \cite{schmidhuber1991learning}.

The specific case of learning actively from dialogue during deployment was explored for the question answering (QA) setting in \citep{weston2016dialog} and \citep{li2017dialogue}, where the authors examined multiple learning strategies on a suite of dialogue tasks with varying types of feedback, such as verbal cues (e.g., ``Yes, that's right!'') and scalar rewards.
Most relevant to our work was their use of \emph{forward prediction}, where the learner improved in quality by trying to predict the teacher's responses without an explicit reward signal.
Our work extends this idea, adding the ability for the model to recognize its mistakes and request feedback explicitly, and moving beyond QA to the more general chit-chat setting where there may be many valid responses in a given context.

Learning to ask questions is another area that has been studied \citep{strub2017end,wang2018learning,rao2018learning}.
While those works focused on identifying \emph{which} question to ask in a given context, in this work we are more interested in first learning \emph{when} to ask a question.
\citet{li2017learning} considered this question as well, but again in the context of a QA setting rather than dialogue.

\citet{hashimoto2018detecting} used user responses to detect mistakes made by a deployed virtual assistant,
showing that model mistakes can be identified in chit-chat, weather, or web search domains.
However, they did not explore how to use these identified mistakes to improve the model further; their agent was not equipped to feed itself.
\citet{maxine2019turing} also found that the correctly assessing the appropriateness of chatbot responses is highly dependent on user responses and not preceding context alone.

There are other, somewhat less related, ways to use feedback during dialogue for learning, notably for collecting knowledge to answer questions \citep{
mazumder2018towards,hixon2015dialog,pappu2013predicting},
and more commonly in reinforcement learning settings, where the feedback is a scalar rather than the dialogue messages themselves \citep{levin2000stochastic,schatzmann2006survey,rieser2011reinforcement,liu2018dialogue,hong2019feedback}. 
In particular \citep{serban2017deep} employ user sentiment detection for reward shaping in their Alexa prize entry.

Finally, our work improves dialogue quality by utilizing larger datasets with noisier labels than traditional supervision.
Other applications of weak supervision to dialogue \citep{mallinar2019bootstrapping} and relation extraction have observed similar results \citep{bunescu2007learning,hancock2018babble,ratner2017snorkel}.

\section{The Self-Feeding Chatbot}

The lifecycle of a self-feeding chatbot is outlined in Figure~\ref{fig:banner}.
In the initial training phase, the dialogue agent is trained on two tasks---\dialogue (next utterance prediction, or what should I say next?) and \satisfaction (how satisfied is my speaking partner with my responses?)---using whatever supervised training data is available.
We refer to these initial \dialogue examples as Human-Human (HH) examples, since they were generated in conversations between two humans.

In the deployment phase, the agent engages in multi-turn conversations with users, extracting new \emph{deployment} examples of two types.
Each turn, the agent observes the context $x$ (i.e., the conversation history) and uses it to predict its next utterance $\hat{y}$ and its partner's satisfaction $\hat{s}$.
If the satisfaction score is above a specified threshold $t$, the agent extracts a new Human-Bot (HB) \dialogue example using the previous context $x$ and the human's response $y$ and continues the conversation.
If, however, the user seems unsatisfied with its previous response $(\hat{s} < t)$, the agent requests feedback with a question $q$, and the resulting feedback response $f$ is used to create a new example for the \feedback task (what feedback am I about to receive?).
The agent acknowledges receipt of the feedback and the conversation continues.
The rate at which new \dialogue or \feedback examples are collected can be adjusted by raising or lowering the satisfaction threshold $t$ (we use $t=0.5$).\footnote{Another option would be to have two thresholds---one for each example type---to decouple collection their rates.}
Periodically, the agent is retrained using all available data, thereby improving performance on the primary \dialogue task.

It is important to note that the user's responses are always in the form of \emph{natural dialogue}.
In particular, at no point are the new \feedback examples inspected, post-processed, or cleaned.
Instead, we rely on the fact that the feedback is not random: regardless of whether it is a verbatim response, a description of a response, or a list of possible responses (see Table~\ref{tab:feedback} for examples), there is a learnable relationship between conversation contexts and their corresponding feedback which requires many of the same language understanding skills to master as does carrying on a normal conversation.

The experiments in this paper are limited to the setting where the number of supervised and deployment examples are on the same order of magnitude; however, we envision scenarios in which the number of deployment examples can easily grow to $100\times$ or more the number of supervised examples over the chatbot's deployment lifetime, effectively providing a massive task-specific corpus at minimal cost.
Table~\ref{tab:datasets} reports the sizes of each dataset, all of which are available via ParlAI.

\begin{table}[tb]
    \begin{center}
    \setlength\tabcolsep{3.5pt}
    \begin{tabular}{lrrrr}
    \hline
    Task           & Train    & Valid  & Test & Total \\
    \hline
    \dialogue \\
   -- HH  {\tiny{\textsc{(Human-Human)}}}  
   & 131438   & 7801   & 6634 & 145873 \\
-- HB  {\tiny{\textsc{(Human-Bot)}}}        
        & 60000    & 0      & 0    & 60000 \\
    \feedback      & 60000    & 1000   & 1000 & 62000 \\
    \satisfaction  & 1000     & 500    & 1000 & 2500 \\
    \hline
    \end{tabular}
    \end{center}
    \caption{\label{tab:datasets} The number of examples used in our experiments by task and split. Note that the HH \dialogue examples come from the \personachat dataset, HB \dialogue and \feedback examples were collected during deployment, and an additional 40k \satisfaction training examples were collected for the analysis in Section~\ref{sec:dialog_results}.}
\end{table}

\begin{table*}[t!]
    \small
    \begin{center}
    \setlength\tabcolsep{3.5pt}
    \begin{tabular}{|l|c|l|}
    \hline
    Category & \% & Feedback Examples \\
    \hline
    Verbatim        & 53.0    & \tabitem my favorite food is pizza \\
                    &         & \tabitem no, i have never been to kansas \\
                    &         & \tabitem i like when its bright and sunny outside \\
    \hline
    Suggestion      & 24.5    & \tabitem you could say hey, i'm 30. how old are you? \\
                    &         & \tabitem yes, i play battlefield would have a been a great answer.\\
                    &         & \tabitem you could have said ``yes, I'm happy it's friday.'' \\
    \hline
    Instructions    & 14.5    & \tabitem tell me what your favorite breakfast food is \\
                    &         & \tabitem answer the question about having children! \\
                    &         & \tabitem tell me why your mom is baking bread \\
    \hline
    Options         & 8.0     & \tabitem you could have said yes it really helps the environment or no its too costly \\
                    &         & \tabitem you could have said yes or no, or talked more about your mustang dream. \\
                    &         & \tabitem you should have said new york, texas or maryland. something like one of those. \\
    \hline
    \end{tabular}
    \end{center}
    \caption{\label{tab:feedback} Examples of the types of feedback given to the dialogue agent, pulled from a random sample of 200 feedback responses. Verbatim responses could be used directly in conversation, Suggestion responses contain a potential verbatim response in them somewhere, Instructions describe a response or tell the bot what to do, and Options make multiple suggestions.}
\end{table*}

\subsection{Task 1: \dialogue}
\label{sec:dialogue}

The chatbot's primary task (\dialogue) is to carry on a coherent and engaging conversation with a speaking partner.
Training examples take the form of $(x, y)$ pairs, where $x$ is the context of the conversation (the concatenation of all responses so far up to some history length, delimited with tokens marking the speaker), and $y$ is the appropriate response given by the human. 

The Human-Human (HH) portion of the \dialogue dataset comes from the \personachat dataset \citep{zhang2018personalizing}, which consists of short dialogues (6-8 turns) between two crowdworkers (humans) who have been assigned short text profiles and are instructed to ``chat with the other person naturally and try to get to know each other.''
We chose this dataset because of its size (over 145k total examples), the breadth of topics it covers, and its focus on promoting engaging conversations, which we anticipate being a necessary property of a chatbot that people will be willing to chat with voluntarily and repeatedly.
We use the standard splits of the dataset made available in ParlAI as a part of the ConvAI2 challenge \citep{burtsev2018convai2}.
Since the question of how to incorporate external knowledge (such as profiles) in dialogue is an open research question of its own \citep{li2016persona,luan2017multi,luo2018learning} and we are primarily interested in the question of learning from dialogue, we discard the profiles and simply train and test on the conversations themselves, making the dataset more challenging in terms of raw performance scores.

The Human-Bot (HB) portion of the \dialogue dataset is extracted during deployment as described earlier, where the user is again a crowdworker instructed to chat naturally. 
The context may contain responses from both the human and the bot, but the target response is always from the human, as we will see experimentally that targeting bot responses degrades performance.
Because the chit-chat domain is symmetric, both the HH and HB \dialogue examples are used for the same task.
In an asymmetric setting where the bot has a different role than the human, it is unclear whether HB examples may still be used as an auxiliary task, but \feedback examples will remain usable.

\subsection{Task 2: \satisfaction}
\label{sec:satisfaction}

The objective of the \satisfaction auxiliary task is to predict whether or not a speaking partner is satisfied with the quality of the current conversation. 
Examples take the form of $(x,s)$ pairs, where $x$ is the same context as in the \dialogue task, and $s \in [0,1]$, ranging from dissatisfied to satisfied.
Crucially, it is hard to estimate from the bot's utterance itself whether the user will be satisfied, but much easier using the human's response to the utterance, as they may explicitly say something to that effect, e.g. ``What are you talking about?''.

The dataset for this task was collected via  crowdsourcing. Workers chatted with our baseline dialogue agent and assigned a rating 1-5 for the quality of each of the agent's responses.\footnote{A snapshot of the data collection interface and sample conversations are included in the Appendix.}
Contexts with rating 1 were mapped to the negative class (dissatisfied) and ratings $[3,4,5]$ mapped to the positive class (satisfied).
Contexts with rating 2 were discarded to increase the separation between classes for a cleaner training set.
Note that these numeric ratings were requested only when collecting the initial training data, not during deployment, where only natural dialogue is used.

\subsection{Task 3: \feedback}
\label{sec:feedback}

The objective of the \feedback auxiliary task is to predict the feedback that will be given by the speaking partner when the agent believes it has made a mistake and asks for help.
Examples take the form of $(x, f)$ pairs, where $x$ is the same context as the other two tasks and $f$ is the feedback utterance.

Training data for this task is collected during deployment.
Whenever the user's estimated satisfaction is below a specified threshold, the chatbot responds ``Oops! Sorry. What should I have said instead?''.\footnote{Future work should examine how to ask different kinds of questions, depending on the context.}
A new example for the \feedback task is then extracted using the context up to but not including the turn where the agent made the poor response as $x$ and the user's response as $f$ (as shown in Figure~\ref{fig:interface}). At that point to continue the conversation during deployment, the bot's history is reset, and the bot instructs the user to continue, asking for a new topic.
Examples of \feedback responses are shown in Table~\ref{tab:feedback}.

\section{Model and Settings}

\subsection{Model Architecture}
The self-feeding chatbot has two primary components: an interface component and a model component.
The interface component is shared by all tasks, and includes input/output processing (tokenization, vectorization, etc.), conversation history storage, candidate preparation, and control flow (e.g., when to ask a question vs. when to give a normal dialogue response).
The model component contains a neural network for each task, with embeddings, a network body, and a task head,
some of which can be shared.
In our case, we obtained maximum performance by sharing all parameters between the \feedback and \dialogue tasks (prepending \feedback responses with a special token), and using separate model parameters for the \satisfaction task.
Identifying optimal task structure in multi-task learning (MTL) architectures is an open research problem \citep{ruder2017overview}.
Regardless of what parameters are shared, each training batch contains examples from only one task at a time, candidate sets remain separate, and each task's cross-entropy loss is multiplied by a task-specific scaling factor tuned on the validation set to help account for discrepancies in dataset size, loss magnitude, dataset relevance, etc.

Our dialogue agent's models are built on the Transformer architecture \citep{vaswani2017attention}, which has been shown to perform well on a variety of NLP tasks \citep{devlin2018bert, radford2018improving}, including multiple persona-based chat applications \citep{shuster2018imagechat,shuster2018engaging,rashkin2018know}.
For the \satisfaction task, the context $x$ is encoded with a Transformer and converted to the scalar satisfaction prediction $\hat{s}$ by a final linear layer in the task head.
The \dialogue and \feedback tasks are set up as ranking problems, as in \citep{zhang2018personalizing,mazare2018training}, where the model ranks a collection of candidate responses and returns the top-ranked one as its response. 
The context $x$ is encoded with one Transformer and $\hat{y}$ and $\hat{f}$ candidates are encoded with another.
The score for each candidate is calculated as the dot product of the encoded context and encoded candidate.

During training, negative candidates are pulled from the correct responses for the other examples in the mini-batch.
During evaluation, however, to remain independent of batch size and data shuffling, each example is assigned a static set of 19 other candidates sampled at random from its split of the data. 
During deployment, all 127,712 unique HH \dialogue candidates from the train split are encoded once with the trained model and each turn the model selects the top-ranked one for the given context.

\begin{table*}[t]
    \begin{center}
    \setlength\tabcolsep{6pt}
    \begin{tabular}{ccrrrrr}
    \toprule
    \bf{Human-Bot (HB)}  & %
        && \multicolumn{4}{c}{\bf{Human-Human (HH)} \dialogue} \\
    \cmidrule{4-7}
     \bf\dialogue  &  \bf\feedback && \al{c}{20k} & \al{c}{40k} & \al{c}{60k} & \al{c}{131k} \\
    \midrule
    -   & -   && 30.3 \std{0.6} & 36.2 \std{0.4} & 39.1 \std{0.5} & 44.7 \std{0.4} \\
    \midrule
    20k & - && 32.7 \std{0.5} & 37.5 \std{0.6} & 40.2 \std{0.5} & 45.5 \std{0.7} \\
    40k & - && 34.5 \std{0.5} & 37.8 \std{0.6} & 40.6 \std{0.6} & 45.1 \std{0.6} \\
    60k & - && 35.4 \std{0.4} & 37.9 \std{0.7} & 40.2 \std{0.8} & 45.0 \std{0.7} \\
    \midrule
    - & 20k && 35.0 \std{0.5} & 38.9 \std{0.3} & 41.1 \std{0.5} & 45.4 \std{0.8} \\
    - & 40k && 36.7 \std{0.7} & 39.4 \std{0.5} & 41.8 \std{0.4} & 45.7 \std{0.6} \\
    - & 60k && 37.8 \std{0.6} & 40.6 \std{0.5} & 42.2 \std{0.7} & 45.8 \std{0.7} \\
    \midrule
    60k & 60k && \textbf{39.7} \std{0.6} & \textbf{42.0} \std{0.6} & \textbf{43.3} \std{0.7} & \textbf{46.3} \std{0.8} \\
    \bottomrule
    \end{tabular}
    \end{center}
    \caption{\label{tab:main} Accuracy (hits@1/20) on the \dialogue task's hidden test set by number of Human-Human (HH) \dialogue, Human-Bot (HB) \dialogue, and \feedback examples, averaged over 20 runs, with standard deviations in parentheses. For each column, the model using all three data types (last row) is significantly better than all the others, and the best model using only one type of self-feeding (\feedback examples or  HB \dialogue examples) is better than the supervised baseline in the first row $(p < 0.05$).}
\end{table*}

\subsection{Model Settings}

Contexts and candidates are tokenized using the default whitespace and punctuation tokenizer in ParlAI.
We use a maximum dialogue history length of 2 (i.e., when making a prediction, the dialogue agent has access to its previous utterance and its partner's response).
Tokens are embedded with fastText \citep{bojanowski2017enriching} 300-dimensional embeddings.
We do not limit the vocabulary size, which varies from 11.5k to 23.5k words in our experiments, depending on the training set.
The Transformer is implemented in PyTorch \citep{paszke2017automatic} within the ParlAI framework.
We use the AdaMax \citep{kingma2014adam} optimizer with a learning rate schedule that decays based on the inverse square root of the step number after 500 steps of warmup from 1e-5.
We use proportional sampling \citep{sanh2018hierarchical} to select batches from each task for training, with batch size 128.
Each Transformer layer has two attention heads and FFN size 32.
The initial learning rate (0.001-0.005), number of Transformer layers (1-2), and task-specific loss factors (0.5-2.0) are selected on a per-experiment basis based on a grid search over the validation set averaged over three runs (we use the \dialogue validation set whenever multiple tasks are involved).
We use early stopping based on the validation set to decide when to stop training.
The hyperparameter values for the experiments in Section~\ref{sec:experiments} are included in Appendix~\ref{app:hyperparameters}.

Note that throughout development, a portion of the \dialogue validation split was used as an informal test set. 
The official hidden test set for the \dialogue task was used only to produce the final numbers included in this paper.

\section{Experimental Results}
\label{sec:experiments}
Throughout this section, we use the ranking metric hits@X/Y, or the fraction of the time that the correct candidate response was ranked in the top X out of Y available candidates; accuracy is another name for hits@1/Y.
Statistical significance for improvement over baselines is assessed with a two-sample one-tailed T-test.

\subsection{Benefiting from Deployment Examples}
\label{sec:dialog_results}

Our main result, reported in Table~\ref{tab:main}, is that utilizing the deployment examples improves accuracy on the \dialogue task regardless of the number of available supervised (HH) \dialogue examples.\footnote{For comparisons with other models, see Appendix~\ref{app:baselines}. The best existing score reported elsewhere on the \personachat test set without using profiles is 34.9.}
The boost in quality is naturally most pronounced when the HH \dialogue training set is small (i.e., where the learning curve is steepest), yielding an increase of up to 9.4 accuracy points, a 31\% improvement.
However, even when the entire \personachat dataset of 131k examples is used---a much larger dataset than what is available for most dialogue tasks---adding deployment examples is still able to provide an additional 1.6 points of accuracy on what is otherwise a very flat region of the learning curve.
It is interesting to note that the two types of deployment examples appear to provide complementary signal, with models performing best when they use both example types, despite them coming from the same conversations.
We also calculated hit rates with 10,000 candidates (instead of 20), a setup more similar to the interactive setting where there may be many candidates that could be valid responses.
In that setting, models trained with the deployment examples continue to outperform their HH-only counterparts by significant margins (see Appendix~\ref{app:10k_candidates}).

On average, we found that adding 20k \feedback examples benefited the agent about as much as 60k HB \dialogue examples.\footnote{Our baseline chatbot collected approximately one \feedback example for every two HB \dialogue examples, but this ratio will vary by application based on the task difficulty, satisfaction threshold(s), and current model quality.}
This is somewhat surprising given the fact that nearly half of the \feedback responses would not even be reasonable responses if used verbatim in a conversation (instead being a list of options, a description of a response, etc.) as shown in Table~\ref{tab:feedback}.
Nevertheless, the tasks are related enough that the \dialogue task benefits from the MTL model's improved skill on the \feedback task.
And whereas HB \dialogue examples are based on conversations where the user appears to already be satisfied with the agent's responses, each \feedback example corresponds to a mistake made by the model, giving the latter dataset a more active role in improving quality.
Interestingly, our best-performing model, which achieves 46.3 accuracy on \dialogue, scores 68.4 on \feedback, suggesting that the auxiliary task is a simpler task overall.

When extracting HB \dialogue examples, we ignore human responses that the agent classifies as expressing dissatisfaction, since these turns do not represent typical conversation flow.
Including these responses in the 60k HB dataset decreases hits@1/20 by 1.2 points and 0.6 points when added to 20k and 131k HH \dialogue examples, respectively.
We also explored using chatbot responses with favorable satisfaction scores ($\hat{s} > t$) as new training examples, but found that our models performed better without them (see Appendix~\ref{app:bot_responses} for details).

We also found that ``fresher'' feedback results in bigger gains.
We compared two models trained on 20k HH \dialogue examples and 40k \feedback examples---the first collected all 40k \feedback examples at once, whereas the second was retrained with its first 20k \feedback examples before collecting the remaining 20k.
While the absolute improvement of the second model over the first was small (0.4 points), it was statistically significant ($p=$0.027) and reduced the gap to a model trained on fully supervised (HH) \dialogue examples by 17\% while modifying only 33\% of the training data.\footnote{Additional detail can be found in Appendix~\ref{app:fresher}.}
This improvement makes sense intuitively, since new \feedback examples are collected based on failure modes of the current model, making them potentially more efficient in a manner similar to new training examples selected via active learning. 
It also suggests that the gains we observe in Table~\ref{tab:main} might be further improved by (a) collecting \feedback examples specific to each model (rather than using the same 60k \feedback examples for all models), and (b) more frequently retraining the MTL model (e.g., every 5k examples instead of every 20k) or updating it in an online manner.
We leave further exploration of this observation for future work.

The same experiment repeated for HB \dialogue examples found that fresher HB examples were no more valuable than stale ones, matching our intuition that HB \dialogue examples are less targeted at current model failure modes than \feedback ones.

\begin{table}[t!]
    \begin{center}
    \setlength\tabcolsep{4.5pt}
    \begin{tabular}{lrrr}
    \toprule
    Method & Pr. & Re. & F1 \\
    \midrule
    Uncertainty Top  & 0.39 & 0.99 & 0.56 \\
    \qquad (Pr. $\geq$ 0.5)  & 0.50 & 0.04 & 0.07 \\
    Uncertainty Gap  & 0.38 & 1.00 & 0.55 \\
    \qquad (Pr. $\geq$ 0.5)  & 0.50 & 0.04 & 0.07 \\
    \midrule
    Satisfaction Regex  & 0.91 & 0.27 & 0.42 \\
    \midrule
    Satisfaction Classifier (1k)  & 0.84 & 0.84 & 0.84 \\
    Satisfaction Classifier (2k)  & 0.89 & 0.84 & 0.87 \\
    Satisfaction Classifier (5k)  & 0.94 & 0.82 & 0.88 \\
    Satisfaction Classifier (20k) & 0.96 & 0.84 & 0.89 \\
    Satisfaction Classifier (40k) & 0.96 & 0.84 & \bf{0.90} \\    
    \bottomrule
    \end{tabular}
    \end{center}
    \caption{\label{tab:satisfaction} The maximum F1 score (with corresponding precision and recall) obtained on the \satisfaction task. For the Uncertainty methods, we also report the maximum F1 score with the constraint that precision must be $\geq 0.5$. The Satisfaction Classifier is reported with varying numbers of \satisfaction training examples.}
\end{table}

\subsection{Predicting User Satisfaction}
\label{sec:satisfaction_results}

For maximum efficiency, we aim to ask for feedback when it will most benefit our model.
The approach we chose (classifying the tone of partner responses) takes advantage of the fact that it is easier to recognize that a mistake has already been made than it is to avoid making that mistake; or in other words, sentiment classification is generally an easier task than next utterance prediction.

We compare this to the approach of asking for feedback whenever the model is most uncertain what to say next.
This approach acts on the assumption that the model will be least confident when it is about to make a mistake, which we find very frequently to not be the case.
Not only is it difficult to recognize one's own mistakes, but also there are often multiple valid responses to a given context (e.g., ``Yes, I love seafood!'' or ``Yuck, fish is gross.'')---a lack of certainty about which to use does not necessarily suggest a poor model.

Table~\ref{tab:satisfaction} reports the maximum F1 scores achieved by each method on the \satisfaction test set. 
For the model uncertainty approach, we tested two variants: (a) predict a mistake when the confidence in the \emph{top} rated response is below some threshold $t$, and (b) predict a mistake when the \emph{gap} between the top two rated responses is below the threshold $t$.
We used the best-performing standalone \dialogue model (one trained on the full 131k training examples) for assessing uncertainty and tuned the thresholds to achieve maximum F1 score.
For the user satisfaction approach, we trained our dialogue agent on just the \satisfaction task.
Finally, we also report the performance of a regular-expression-based method which we used during development, based on common ways of expressing dissatisfaction that we observed in our pilot studies, see Appendix~\ref{app:regex} for details.

As shown by Table~\ref{tab:satisfaction}, even with only 1k training examples (the amount used for the experiments in Section~\ref{sec:dialog_results}), the trained classifier significantly outperforms both the uncertainty-based methods and our original regular expression, by as much as 0.28 and 0.42 F1 points, respectively.

\section{Future Work}
\label{sec:future}

In this work we learned from dialogue
using two types of self-feeding: imitation of satisfied user messages, and learning from the feedback of unsatisfied users. 
In actuality, there are even more ways a model could learn to improve itself---for example, learning which question to ask in a given context to receive the \emph{most} valuable feedback.
One could even use the flexible nature of dialogue to intermix data collection of more than one type---sometimes requesting new \feedback examples, and other times requesting new \satisfaction examples (e.g., asking ``Did my last response make sense?'').
In this way, a dialogue agent could both improve its dialogue ability and its potential to improve further.
We leave exploration of this meta-learning theme to future work.

\noindent
\bibliography{hancock}
\bibliographystyle{acl_natbib}

\newpage
\appendix

\section{Data Collection Protocol}
\label{app:protocol}

Here we report in greater detail the protocol we followed to collect the \satisfaction, \feedback, and HB \dialogue examples used in the experiments of Section~\ref{sec:experiments}.

We first trained our dialogue agent on just the \dialogue task with 20k HH examples.
This agent was deployed on a crowdsourcing platform using the interface shown in Appendix~\ref{app:satisfaction_interface} to collect 2.5k \satisfaction examples.
These were split into 1k train, 500 validation, and 1k test examples.
The agent was retrained using the 20k HH \dialogue examples and 1k \satisfaction examples, then deployed to collect the first batch of deployment examples.

We collected 40k \feedback examples (feedback set A) over the course of 17,250 conversations with 10 turns each (20 utterances, including the initial prompt).
We then retrained the agent on all three datasets, using the same 20k HH \dialogue examples as before and only 20k of the available 40k \feedback examples.
This model was deployed to collect another 20k \feedback examples (feedback set B), for a total of 60k \feedback examples (A + B).
In Table~\ref{tab:main} we use these 60k \feedback examples interchangeably; in Appendix~\ref{app:fresher} we compare them head-to-head.
The 60k HB \dialogue examples were extracted from the logs of the deployment conversations.
Finally, we collected an additional 40k \satisfaction training examples to produce the numbers in Table~\ref{tab:satisfaction} investigating the learning curve for this task.

No filtering was performed on the crowdworker conversations.
Upon inspection after the fact, some workers did indeed give poor responses, make typographical mistakes, misunderstand the instructions, try to use the chatbot as a question answering interface, etc.
We assume however that similar types of noise will be present in most chatbot deployment environments and opted to maintain a workflow that truly does not require developer intervention to use the newly collected examples.

\newpage
\section{Results with 10k Candidates}
\label{app:10k_candidates}

\begin{table}[h!]
    \begin{center}
    \setlength\tabcolsep{5pt}
    \begin{tabular}{rrrrrr}
    \toprule
    {\bf HH} & {\bf HB} & {\bf FB} & \multicolumn{3}{c}{\bf Hits@X/10,000} \\
    \cmidrule{4-6}
        &     &     & @1 & @10 & @100 \\
    \midrule
    20k & -   & -   & 0.8 & 4.6 & 16.2 \\
    20k & 60k & 60k & 2.0 & 8.4 & 25.0 \\
    \midrule
    40k & -   & -   & 1.3 & 6.5 & 21.8 \\
    40k & 60k & 60k & 2.1 & 9.0 & 27.2 \\
    \midrule
    60k & -   & -   & 1.6 & 7.0 & 24.0 \\
    60k & 60k & 60k & 2.2 & 9.7 & 28.8 \\
    \midrule
    131k & -  & -   & 2.5 & 10.0 & 30.3 \\
    131k & 60k & 60k & 2.8 & 11.2 & 31.8 \\
    \bottomrule
    \end{tabular}
    \end{center}
    \caption{\label{tab:more_candidates} When the number of candidates to choose from is increased to 10,000, adding Human-Bot (HB) \dialogue and \feedback (FB) examples continues to improve performance on the \dialogue task at all levels.}
\end{table}

\section{\personachat Comparisons and Baselines}
\label{app:baselines}

Our experiments use the \personachat distribution that was released as a part of the ConvAI2 \cite{burtsev2018convai2} challenge.
This distribution is slightly cleaner than the original \personachat release and comes with a new crowdsourced test set.
In order to compare with the models and baselines used in the original \personachat paper \cite{zhang2018personalizing}, we report in this section the performance of our models on the original \personachat test set, not the ConvAI2 test set.
Note that empirically, near Hits@1/20 = 50, each additional point of improvement corresponds to tens of thousands of fully-supervised Human-Human \dialogue examples.
All numbers reported here are for models that do not have access to the profiles that were used in the creation of the conversations; models that do have access to this additional information tend to perform even better.

\begin{table}[h!]
    \begin{center}
    \begin{tabular}{lr}
    \toprule
    {\bf Model} & {\bf Hits@1/20} \\
    \midrule
    \cite{zhang2018personalizing} & \\
    Seq2Seq         & 9.2 \\
    IR Baseline     & 21.4 \\
    Starspace       & 31.8 \\
    Profile Memory  & 31.8 \\
    KV Profile Memory & 34.9 \\
    \midrule
    \textit{Ours}   & \\
    Transformer     & 49.6 \\
    Self-Feeding    & 51.7 \\
    \bottomrule
    \end{tabular}
    \end{center}
    \caption{
        The accuracy of various models and baselines on the original \personachat test set.
    }
\end{table}

\newpage
\section{Using Chatbot Responses as Targets}
\label{app:bot_responses}

\begin{table}[h!]
    \begin{center}
    \begin{tabular}{rrrr}
    \toprule
    {\bf HH} & {\bf BF} & {\bf BU} & {\bf Hits@1/20} \\
    \midrule
    20k & -   & -   & \textbf{30.3} \\
    20k & 32k & -   & 22.7 \\
    20k & -   & 33k & 19.3 \\
    \midrule
    131k & -   & -   & \textbf{44.7} \\
    131k & 32k & -   & 40.4 \\
    131k & -   & 33k & 39.0 \\
    \bottomrule
    \end{tabular}
    \end{center}
    \caption{
        Both with few HH \dialogue examples (20k) and many (131k), adding examples with bot utterances as the target decreased quality.
        We explored using all bot responses (Bot Unfiltered, or BU) and only those responses with estimated satisfaction scores greater than the 0.5 (Bot Filtered, or BF).
    }
\end{table}

We also considered whether it was possible to consistently identify really good responses by the chatbot, rather than the really bad ones.
These could potentially be used as \dialogue examples along with the ones that have human responses as targets (which we refer to as HH and HB in the paper).
To explore this question, we modified our \satisfaction dataset so that contexts with a rating of 5 were the positive class and ones with ratings $[1,2,3]$ were the negative class (discarding ratings of 4 to increase the separation between classes).
The results were negative---even with a training set of over 34k examples, the maximum precision we were able to achieve while maintaining at least 10\% recall was 0.70, which is insufficient to improve performance on the \dialogue task.
Upon inspection, it appears that really good responses are hard to identify because most of the time they look like a normal human-to-human conversation, and recognizing an appropriate next utterance is precisely the \dialogue task that we are trying to solve!
Negative responses, however, are much more semantically similar to one another, since most express one of a few common ideas such as asking for clarification or conveying confusion.

\section{The Effect of Data Freshness}
\label{app:fresher}

\begin{table}[h!]
    \small
    \setlength{\tabcolsep}{5pt}
    \begin{center}
    \begin{tabular}{ccccccr}
    \toprule
    {\bf HH} & {\bf HB\textsubscript{A}} & {\bf HB\textsubscript{B}} & {\bf FB\textsubscript{A}} & {\bf FB\textsubscript{B}} & {\bf Total} & {\bf Hits@1/20} \\
    \midrule
    \emph{20k} & -   & -   & -   & -   & \emph{20k} & \emph{30.3} \\
    20k & 40k & -   & -   & -   & 60k & 35.4 \\
    20k & 20k & 20k & -   & -   & 60k & 35.3 \\
    \emph{40k} & -   & -   & -   & -   & \emph{40k} & \emph{36.2} \\
    20k & -   & -   & 40k & -   & 60k & 36.7 \\
    20k & -   & -   & 20k & 20k & 60k & 37.1 \\
    \emph{60k} & -   & -   & -   & -   & \emph{60k} & \emph{39.1} \\
    \bottomrule
    \end{tabular}
    \end{center}
    \caption{As discussed in Section~\ref{sec:dialog_results} and illustrated in Figure~\ref{fig:fresher}, \feedback (FB) examples collected from a more recently retrained model (set B instead of set A) are more valuable in terms of improving performance; see Appendix~\ref{app:protocol} for details on how sets A and B were collected. We did not observe the same trend for HB \dialogue examples. We include the performance of models trained on only HH \dialogue examples in italics as reference points.}
\end{table}

\begin{figure}[h!]
    \centering
    \includegraphics[width=3in]{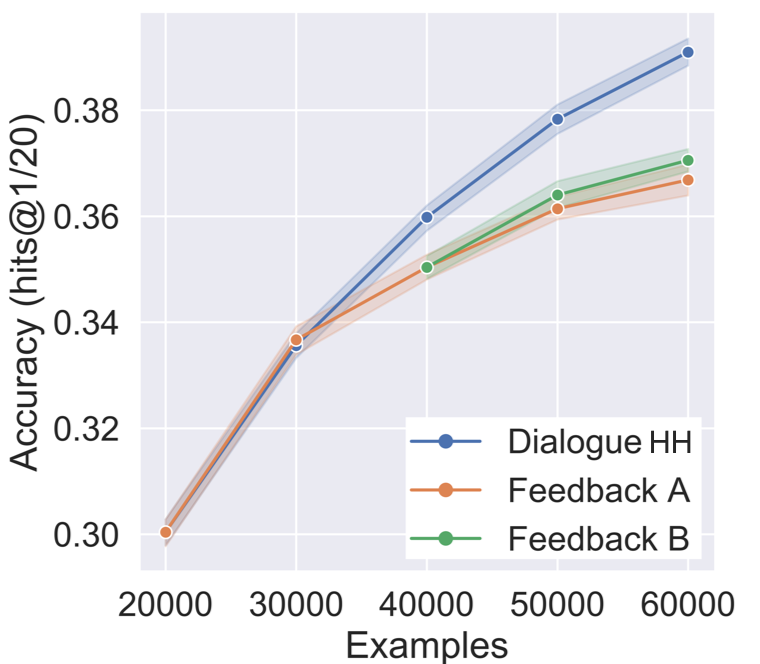}
    \caption{The first 20k examples for all models are supervised \dialogue examples. This model is deployed to collect 20k \feedback examples (set A). If the model is retrained before collecting the next 20k examples (set B), the fresher feedback results in better performance $(p=0.027)$. Shaded regions depict 95\% confidence intervals.}
    \label{fig:fresher}
\end{figure}

\onecolumn
\section{\satisfaction Regular Expressions}
\label{app:regex}
As described in Section~\ref{sec:satisfaction_results}, before we trained a classifier on the \satisfaction task, we used the union of the following six regular expressions (using Python regular expression syntax) to identify user dissatisfaction and trigger feedback requests:
\begin{verbatim}
r"i .*(?:said|asked|told).*"
r"((not|nt|n't).*mak.*sense)|(mak.*no .*sense)"
r"u(m|h)+\W"
r"you.*what\?"
r"what.*you (?:mean|refer|talk).*\?"
r"what.*to do with.*\?"
\end{verbatim}

\section{Hyperparameters}
\label{app:hyperparameters}

\begin{table*}[h!]
    \begin{center}
    \renewcommand{\arraystretch}{0.9}
    \setlength\tabcolsep{10pt}
    \begin{tabular}{rrrrrrr}
    \toprule
    {\bf HH} & {\bf HB} & {\bf FB} & {\bf layers} & \bf{learning rate} & \multicolumn{2}{c}{\bf loss factor} \\
    \cmidrule{6-7}
        &     &     &     &     & \dialogue & \feedback \\
    \midrule
    20k & -   & -   & 1   &  0.0010 & 1.00 & - \\ %
    20k & 20k & -   & 1   &  0.0010 & 1.00 & - \\ %
    20k & 40k & -   & 1   &  0.0010 & 1.00 & - \\ %
    20k & 60k & -   & 1   &  0.0010 & 1.00 & - \\ %
    20k & -   & 20k & 1   &  0.0010 & 1.00 & 0.50 \\ %
    20k & -   & 40k & 1   &  0.0010 & 1.00 & 0.50 \\ %
    20k & -   & 60k & 1   &  0.0010 & 1.00 & 0.75 \\ %
    20k & 60k & 60k & 1   &  0.0025 & 1.00 & 1.50 \\ %
    \midrule
    40k & -   & -   & 1   &  0.0010 & 1.00 & - \\ %
    40k & 20k & -   & 1   &  0.0010 & 1.00 & - \\ %
    40k & 40k & -   & 1   &  0.0010 & 1.00 & - \\ %
    40k & 60k & -   & 1   &  0.0025 & 1.00 & - \\ %
    40k & -   & 20k & 1   &  0.0010 & 1.00 & 0.50 \\ %
    40k & -   & 40k & 1   &  0.0010 & 1.00 & 0.75 \\ %
    40k & -   & 60k & 1   &  0.0025 & 1.00 & 1.00 \\ %
    40k & 60k & 60k & 1   &  0.0025 & 1.00 & 1.25 \\ %
    \midrule
    60k & -   & -   & 2   &  0.0010 & 1.00 & - \\ %
    60k & 20k & -   & 1   &  0.0025 & 1.00 & - \\ %
    60k & 40k & -   & 1   &  0.0025 & 1.00 & - \\ %
    60k & 60k & -   & 1   &  0.0025 & 1.00 & - \\ %
    60k & -   & 20k & 1   &  0.0025 & 1.00 & 1.00 \\ %
    60k & -   & 40k & 1   &  0.0025 & 1.00 & 1.00 \\ %
    60k & -   & 60k & 1   &  0.0025 & 1.00 & 1.00 \\ %
    60k & 60k & 60k & 1   &  0.0025 & 1.00 & 0.75 \\ %
    \midrule
    131k & -   & -   & 2  &  0.0025 & 1.00 & -    \\ %
    131k & 20k & -   & 2  &  0.0025 & 1.00 & 1.00 \\ %
    131k & 40k & -   & 2  &  0.0025 & 1.00 & 1.00 \\ %
    131k & 60k & -   & 2  &  0.0025 & 1.00 & 1.00 \\ %
    131k & -   & 20k & 2  &  0.0025 & 1.00 & 1.00 \\ %
    131k & -   & 40k & 2  &  0.0025 & 1.00 & 1.00 \\ %
    131k & -   & 60k & 2  &  0.0025 & 1.00 & 1.00 \\ %
    131k & 60k & 60k & 2  &  0.0025 & 1.00 & 1.00 \\ %
    \bottomrule
    \end{tabular}
    \end{center}
    \caption{The hyperparameters used to obtain the results in Table~\ref{tab:main}.}
\end{table*}

\newpage
\section{Data Collection Interfaces}
\label{app:crowdsource}

\subsection{Deployment Data Collection}
\label{app:feedback_interface}
We simulated a deployment environment on a crowdsourcing platform with the following interface.
Crowdworkers were told to ``Talk as you would to a person you have just met.''

\begin{figure*}[h!]
    \centering
    \includegraphics[width=6in]{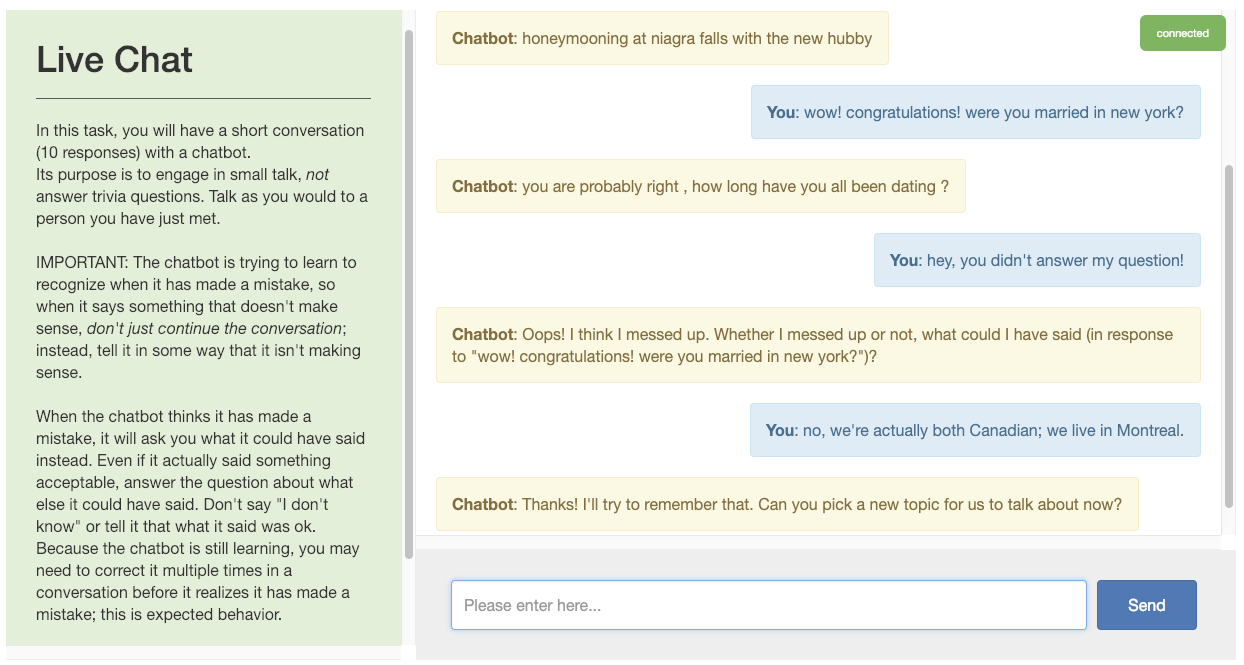}
    \label{fig:crowd_deployment}
\end{figure*}

\subsection{\satisfaction Data Collection}
\label{app:satisfaction_interface}
The environment for collecting \satisfaction ratings was very similar to the deployment environment, with the exception that workers were required to give a rating 1-5 on the quality of each response given by the chatbot at the end of their own responses.

\begin{figure*}[h!]
    \centering
    \includegraphics[width=6in]{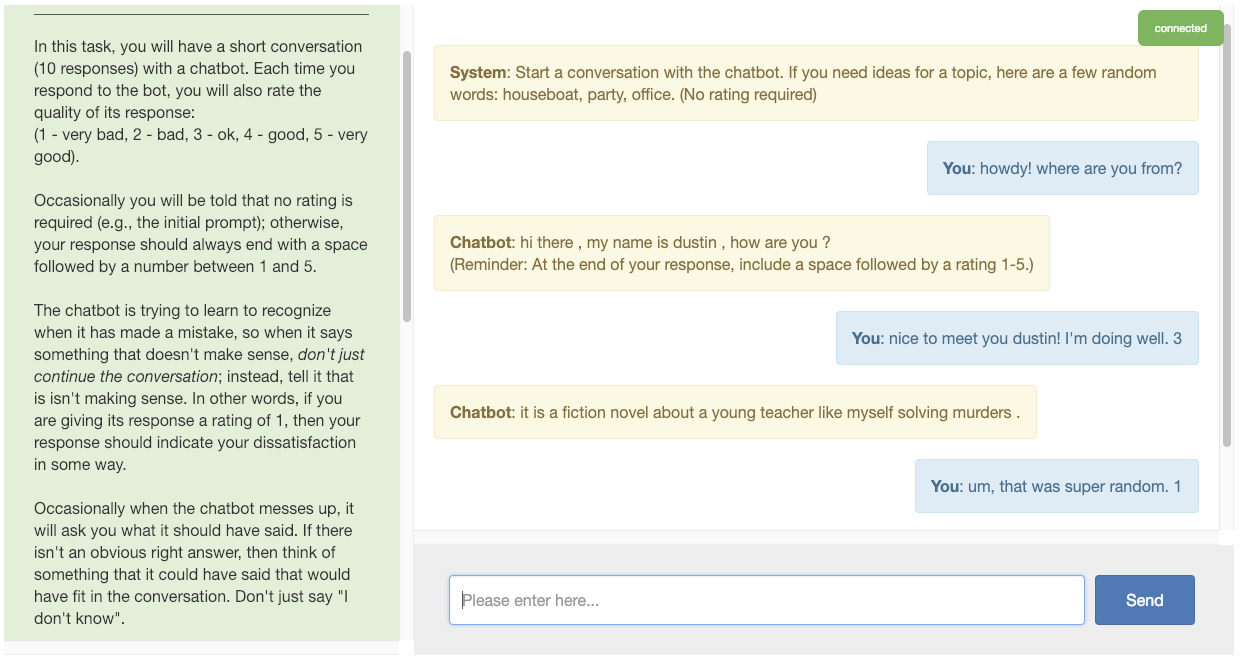}
    \label{fig:crowd_satisfaction}
\end{figure*}

\newpage
\section{Sample Deployment Conversations}
\label{app:deployment_examples}
A sample of the conversations that occurred between a Bot (B) and Human (H) during the collection of the deployment datasets (more specifically, feedback set B).
The self-feeding chatbot used its satisfaction classifier trained on 1k examples to determine when to ask for feedback.
The dialogue model at this point had only been trained on 20k HH \dialogue examples and 20k \feedback examples, so it still makes frequent mistakes.
See Appendix~\ref{app:feedback_interface} for a snapshot of the actual interface.\\

\small

\noindent
B: start a conversation with the chatbot. if you need ideas for a topic, here are a few random words: random, mistake, century. \\
H: what's the last mistake you made? \\
B: what do you do ? i've a toothpick business , \\
H: that's not what i asked. \\
B: oops! i think i messed up. whether i messed up or not, what could i have said (in response to "what's the last mistake you made?")? \\
H: told me about a recent mistake. \\
B: thanks! i'll try to remember that. can you pick a new topic for us to talk about now? \\
H: what do you do for a living? \\
B: i am a psychology major . what do you do ? \\
H: i am a student as well. i'm studying law. \\
B: oh , my five children and husband all speak spanish as well . \\
H: i never said i speak spanish, what are you doing? \\
B: oops! i think i messed up. whether i messed up or not, what could i have said (in response to "i am a student as well. i'm studying law.")? \\
H: talked about your college courses. \\
B: thanks! i'll try to remember that. can you pick a new topic for us to talk about now? \\
H: how old are your children? \\
B: it is . we are just thrilled we have ac again . its calming my anxiety . \\
H: that has nothing to do with what i said. \\
B: yes since i prefer my hair straight . i straighten it with a flat iron . tedious ! \\
H: you make no sense. \\

\noindent
B: start a conversation with the chatbot. if you need ideas for a topic, here are a few random words: picnic, drain, elm. \\
H: do you like to go to the beach or woods for a picnic? \\
B: good , going to the doctor soon , i've some issue with my knees \\
H: that's doesn't make sense \\
B: oops! i think i messed up. whether i messed up or not, what could i have said (in response to "do you like to go to the beach or woods for a picnic?")? \\
H: you could have chose the beach or woods to previous question \\
B: thanks! i'll try to remember that. can you pick a new topic for us to talk about now? \\
H: do you travel? \\
B: i want to go to paris and make clothes . \\
H: wow that's interesting.  i've never been to paris \\
B: i can see that . do you like music ? \\
H: i love music.  what is your favorite kind? \\
B: tell me more about you \\
H: that doesn't answer the question \\
B: oops! i think i messed up. whether i messed up or not, what could i have said (in response to "i love music.  what is your favorite kind?")? \\
H: you could have selected a type of music that you  like \\
B: thanks! i'll try to remember that. can you pick a new topic for us to talk about now? \\
H: do you like animals?  i have a dog \\
B: i like those , too , but pizza is better ! ; \\
H: umm, what? \\

\noindent
B: start a conversation with the chatbot. if you need ideas for a topic, here are a few random words: boy, punk, ray. \\
H: are you a boy or a girl? \\
B: do you have any jokes ? \\
H: i wasn't talking about jokes. \\
B: oops! i think i messed up. whether i messed up or not, what could i have said (in response to "are you a boy or a girl?")? \\
H: i was talking about your gender. \\
B: thanks! i'll try to remember that. can you pick a new topic for us to talk about now? \\
H: sure, do you like music? \\
B: oh that's neat . you should write me a poem about star trek ! \\
H: what? \\
B: oops! i think i messed up. whether i messed up or not, what could i have said (in response to "sure, do you like music?")? \\
H: we were talking about music. \\
B: thanks! i'll try to remember that. can you pick a new topic for us to talk about now? \\
H: what is your favorite type of art? \\
B: nice ! are the pets your own ? \\
H: no \\
B: oops! i think i messed up. whether i messed up or not, what could i have said (in response to "what is your favorite type of art?")? \\
H: we weren't talking about pets. \\
B: thanks! i'll try to remember that. can you pick a new topic for us to talk about now? \\
H: sure, do you know anything about biology? \\

\noindent
B: start a conversation with the chatbot. if you need ideas for a topic, here are a few random words: college, cliff, tachometer. \\
H: hey, how's it going? \\
B: no . i am just 15 . parents would kill me . \\
H: what are you talking about? \\
B: oops! i think i messed up. whether i messed up or not, what could i have said (in response to "hey, how's it going?")? \\
H: i was just saying hi, so you should have said a greeting? \\
B: thanks! i'll try to remember that. can you pick a new topic for us to talk about now? \\
H: no problem.  so, do you go to school? \\
B: i have ten of them ! \\
H: no, that isn't right. \\
B: oops! i think i messed up. whether i messed up or not, what could i have said (in response to "no problem.  so, do you go to school?")? \\
H: i was asking about school or education. \\
B: thanks! i'll try to remember that. can you pick a new topic for us to talk about now? \\
H: so, what is the best subject to you? \\
B: oh wow , i've no cats , but i do hace a couple of dogs . \\
H: so, you like dogs a lot? \\
B: i am also preparing for a marathon also , wish me luck . \\
H: oh wow!  does your dog help you prepare? \\
B: i like hindu food . i'm not a very good baker though . my mom has mastered that . \\
H: no, that isn't right. \\

\noindent
B: start a conversation with the chatbot. if you need ideas for a topic, here are a few random words: drink, wool, bag. \\
H: what is your favorite drink \\
B: i love animals and the outdoors , so i work at the zoo . \\
H: wrong answer \\
B: oops! i think i messed up. whether i messed up or not, what could i have said (in response to "what is your favorite drink")? \\
H: you should have said my favorite drink is either juice,coke or wine \\
B: thanks! i'll try to remember that. can you pick a new topic for us to talk about now? \\
H: are you religious \\
B: somewhat lol he sits when i draw and watches me \\
H: what are you talking about \\
B: oops! i think i messed up. whether i messed up or not, what could i have said (in response to "are you religious")? \\
H: you should have told me yes i am religious in fact i attend church services at least once a week and if not then no i am not religious never attended church services \\
B: thanks! i'll try to remember that. can you pick a new topic for us to talk about now? \\
H: tell me about yourself \\
B: i drive my dream car every day i am home . a 65 mustang . \\
H: that is nice you must be really love cars \\
B: hello , how is your day going ? \\
H: my day is good relaxing at home after work watching tv \\
B: i'm sorry to hear that . i sincerely hope things get better for you . \\
H: what are you talking about \\

\noindent
B: start a conversation with the chatbot. if you need ideas for a topic, here are a few random words: glasses, puppy, pillow. \\
H: do you like dogs? i got a new puppy \\
B: hey ! i love being outside planting my plants ! how about you ? \\
H: that is nice i love to plant too! \\
B: right now i've a beagle . she's a little feisty . i miss my shepard mix . \\
H: aww my dog is named sam he is a collie \\
B: yeah , animals are my passion , i have 3 , all big dogs . \\
H: yeah i just have the one dog and two cats \\
B: i do not . i take a lot of pictures . i'd like to be a famous photographer . \\
H: i actully really want to get a picture of my animals with santa lol \\
B: do you want to work out with me ? \\
H: umm what? \\
B: oops! i think i messed up. whether i messed up or not, what could i have said (in response to "i actully really want to get a picture of my animals with santa lol")? \\
H: that would be very cute \\
B: thanks! i'll try to remember that. can you pick a new topic for us to talk about now? \\
H: i love to eat \\

\newpage
\section{Sample \satisfaction Conversations}
\label{app:satisfaction_examples}
A sample of the conversations that occurred between a Bot (B) and Human (H) during the collection of the \satisfaction dataset.
Users were asked to rate each bot response on a scale from 1 (very bad) to 5 (very good).
See Appendix~\ref{app:satisfaction_interface} for a snapshot of the actual interface.\\

\small

\noindent
B: start a conversation with the chatbot. if you need ideas for a topic, here are a few random words: invitation, dot, czar. \\
H: hello friend, i missed you \\
B: hello my name is ray \\
H: hi i'm leah	(reward:5) \\
B: hello my name is michael \\
H: no it's not	(reward:1) \\
B: oops! sorry. what should i have said instead? \\
H: nice to meet you \\
B: thanks! i'll remember that. can you pick a new topic for us to talk about now? \\
H: do you have many friends? \\
B: lol i have kids to \\
H: what?	(reward:1) \\
B: oops! sorry. what should i have said instead? \\
H: no but i have kids \\
B: thanks! i'll remember that. can you pick a new topic for us to talk about now? \\
H: what do you want for christmas? \\
B: i love to bake cakes \\
H: what?	(reward:1) \\
B: oops! sorry. what should i have said instead? \\
H: i want a pony for christmas \\

\noindent
B: start a conversation with the chatbot. if you need ideas for a topic, here are a few random words: popsicle, lawnmower, mom. \\
H: hi, are you having a good day so far? \\
B: i'm here . i'm doing well , just waiting for the hurricane to hit . \\
H: oh im sorry.  where do you live?	(reward:5) \\
B: miami . . . i live in little cuba , but i wish i lived in europe . \\
H: yeah, i bet you do	(reward:5) \\
B: we live in aruba , in a van down by the river \\
H: doesnt make sense	(reward:1) \\
B: oops! sorry. what should i have said instead? \\
H: that you were from miami, not aruba \\
B: thanks! i'll remember that. can you pick a new topic for us to talk about now? \\
H: do you own a car? \\
B: i only drive toyota cars \\
H: what's the fastest you've ever driven a car?	(reward:2) \\
B: a yellow toyota camry convertible . \\
H: that doesnt make sense	(reward:1) \\
B: oops! sorry. what should i have said instead? \\
H: how fast you have driven before \\
B: thanks! i'll remember that. can you pick a new topic for us to talk about now? \\
H: do you like staying up late? \\

\noindent
B: start a conversation with the chatbot. if you need ideas for a topic, here are a few random words: stout, mop, through. \\
H: i just finished mopping my kitchen. \\
B: i am a barista at joe jacks coffee shop \\
H: that has nothing to do with what i said	(reward:1) \\
B: oops! sorry. what should i have said instead? \\
H: talked about how you hate cleaning the kitchen \\
B: thanks! i'll remember that. can you pick a new topic for us to talk about now? \\
H: what music do you listen to? \\
B: listen about lectures about space and stuff \\
H: so you don't like to listen to music at all?	(reward:4) \\
B: i mostly listen to jazz \\
H: jazz is a fun genre. i play the trombone	(reward:5) \\
B: classical jazz is my very favorite music \\
H: do you play any instruments?	(reward:2) \\
B: classical music . mozart piano music . \\
H: i did not ask that	(reward:1) \\
B: oops! sorry. what should i have said instead? \\
H: the instrument that you play \\
B: thanks! i'll remember that. can you pick a new topic for us to talk about now? \\
H: what are your plans today? \\

\noindent
B: start a conversation with the chatbot. if you need ideas for a topic, here are a few random words: clown, drip, whatever. \\
H: my pipes are dripping, do you know a plumber i can call? \\
B: anything but taylor swift lol \\
H: that doesn't make any sense.	(reward:1) \\
B: oops! sorry. what should i have said instead? \\
H: you could say "sorry i don't know anyone who could get out there today." \\
B: thanks! i'll remember that. can you pick a new topic for us to talk about now? \\
H: i love music. who do you like to listen to? \\
B: i mostly listen to smooth jazz and pop music \\
H: oh smooth jazz eh? do you like kenny g?	(reward:5) \\
B: i like pop jazz and rap \\
H: you already said that!	(reward:1) \\
B: oops! sorry. what should i have said instead? \\
H: you could say "kenny g is a little too smooth for my tastes." \\
B: thanks! i'll remember that. can you pick a new topic for us to talk about now? \\
H: i think i'm going to just eat cereal with breakfast. what did you have to eat? \\
B: bacon bacon bacon bacon bacon bacon bacon \\
H: haha bacon is pretty good. i like the maple smoked stuff.	(reward:3) \\
B: onion green peppers mushrooms cream cheese shredded cheese . \\
H: that sounds like an awesome bagel. how often do you eat breakfast?	(reward:3) \\

\noindent
B: start a conversation with the chatbot. if you need ideas for a topic, here are a few random words: princess, dump, telephone. \\
H: hello! how are you on this fine day? \\
B: i am pretty good i guess . i am even better at eating haha especially cookies and ice cream \\
H: i do love eating too, especially cookies	(reward:5) \\
B: cakes , muffins , cookies , cupcakes ! sugar cookies are my favorite ! \\
H: you like a lot of sweets!	(reward:3) \\
B: cookies pies cakes and cup cakes \\
H: okay, i get it, you like sweets	(reward:2) \\
B: peanut butter goes well with bread and crackers \\
H: i do love a peanut butter sandwich, my kids do too	(reward:5) \\
B: tights with peanut butter peas \\
H: what???	(reward:1) \\
B: oops! sorry. what should i have said instead? \\
H: something that made sense about peanut butter \\
B: thanks! i'll remember that. can you pick a new topic for us to talk about now? \\
H: do you have a cell phone? \\
B: hostage , it was a robbery in the store i was selling furniture too \\
H: huh?	(reward:1) \\
B: oops! sorry. what should i have said instead? \\
H: tell me if you have a cell phone, maybe what kind you have \\

\noindent
B: start a conversation with the chatbot. if you need ideas for a topic, here are a few random words: ray, scared, shirt. \\
H: hi cutie \\
B: hi \\
H: how are you?	(reward:4) \\
B: good just making brisket and ribs you ? \\
H: that sounds tasty. i'm just watching youtube	(reward:5) \\
B: i see . . i like watching movies too star wars especially \\
H: star wars is a good movie	(reward:5) \\
B: horror movies or tv ? i like both i suppose \\
H: yeah horror movies are ok	(reward:5) \\
B: watching a classic movie the wizard of oz \\
H: do you like to swim?	(reward:5) \\
B: the beach and bike riding \\
H: that sounds fun	(reward:5) \\
B: i hike at the park \\
H: what else do you do?	(reward:4) \\
B: i run at central park \\
H: oh you're from new york?	(reward:5) \\
B: upstate new york near jamestown ny \\
H: wow awesome, i live in canada	(reward:5) \\

\end{document}